\documentclass[runningheads]{llncs}

\usepackage{graphicx}
\usepackage{amsmath}
\usepackage{amssymb}
\usepackage{multirow}
\usepackage{array}
\usepackage{pifont}
\usepackage{makecell}
\usepackage{caption}
\usepackage{capt-of}
\usepackage{booktabs}
\usepackage{xcolor}
\usepackage{enumitem}
\usepackage{listings}
\usepackage{subcaption}
\usepackage{placeins}
\usepackage{tabularx}
\usepackage[hidelinks]{hyperref}
\usepackage{orcidlink}

\newcommand{\cmark}{\ding{51}}
\newcommand{\xmark}{\ding{55}}

\newcolumntype{Y}{>{\raggedright\arraybackslash}X}

\begin{document}

\title{Food-R1: A Unified Multi-Task Food Vision-Language Model with Reinforcement Learning}
\titlerunning{Food-R1}

\author{
    Yu Zhu\inst{1}\textsuperscript{*} \and
    Yongkang Li\inst{1}\textsuperscript{*} \and
    Wenjie Zhu\inst{1} \and
    Haoyi Jiang\inst{1} \and
    Wenyu Liu\inst{1} \and
    Wei Yang\inst{1} \and
    Bin Li\inst{1} \and
    Xinggang Wang\inst{1}\textsuperscript{\dag}
}

\authorrunning{Y. Zhu et al.}

\institute{
    Huazhong University of Science and Technology, Wuhan, China \\
    \email{\{zyzyzy, xgwang\}@hust.edu.cn} \\
    \textsuperscript{*}Equal contribution. \quad
    \textsuperscript{\dag}Corresponding author. \\
}

\maketitle

\begin{abstract}
Recent studies have explored Vision-Language Models (VLMs) for food analysis. However, most existing methods rely primarily on supervised fine-tuning (SFT), which often limits reasoning and generalization capabilities. Moreover, high-quality large-scale nutritional annotations remain scarce. To address these issues, we introduce \textbf{CalorieBench-80K}, a large-scale benchmark with curated calorie labels and dietary advice annotations. To the best of our knowledge, it is the first food image benchmark to incorporate Chain-of-Thought (CoT) annotations for calorie reasoning. We also propose \textbf{Food-R1}, a unified food VLM trained in a multi-task learning paradigm to equip the model with broad capabilities. Food-R1 undergoes CoT-based cold-start instruction tuning, followed by reinforcement fine-tuning (RFT) using Group Relative Policy Optimization (GRPO) to improve reasoning and performance. Experiments on CalorieBench-80K and representative benchmarks show that Food-R1 consistently outperforms strong baselines across food-related tasks. The code and model weights are available at \url{https://github.com/hustvl/Food-R1} and \url{https://huggingface.co/collections/zy12123/food-r1}.


\end{abstract}

\section{Introduction}
Food analysis aims to understand and reason over food images, including tasks such as ingredient recognition, recipe generation, and nutrition estimation. Purely visual recognition methods are widely used for dish classification~\cite{bossard2014food101,liu2024canteenfood}, detection~\cite{bolanos2017simultaneousfood}, and segmentation~\cite{ciocca2017foodrecognitiondataset,battini2023segmenteduecfood100}. While effective for dish-level recognition, they often struggle in complex, real-world settings due to fine-grained ingredient ambiguity, occlusion in mixed dishes, and appearance variations induced by diverse cooking styles. Consequently, extending these methods to ingredient-level understanding remains challenging.

Recent advances in Vision--Language Models have enabled new progress in food analysis. With broad world knowledge and multimodal reasoning, VLMs have been applied to food classification~\cite{yin2024foodlmm}, ingredient recognition~\cite{jiao2024rode}, recipe generation~\cite{mohbat2024llavachef}, and nutrition estimation~\cite{tanabe2025reasoningdriven}. Recent studies also explore agentic strategies with retrieval and external tools, such as nutrition database retrieval~\cite{yao2024caloraify} and volume estimation modules~\cite{tanabe2025calorievol}, to provide additional structured knowledge. However, most VLM-based methods still rely primarily on supervised fine-tuning (SFT), limiting reasoning consistency and generalization in real-world scenarios.

\begin{figure}[t]
  \centering
  \includegraphics[width=\textwidth]{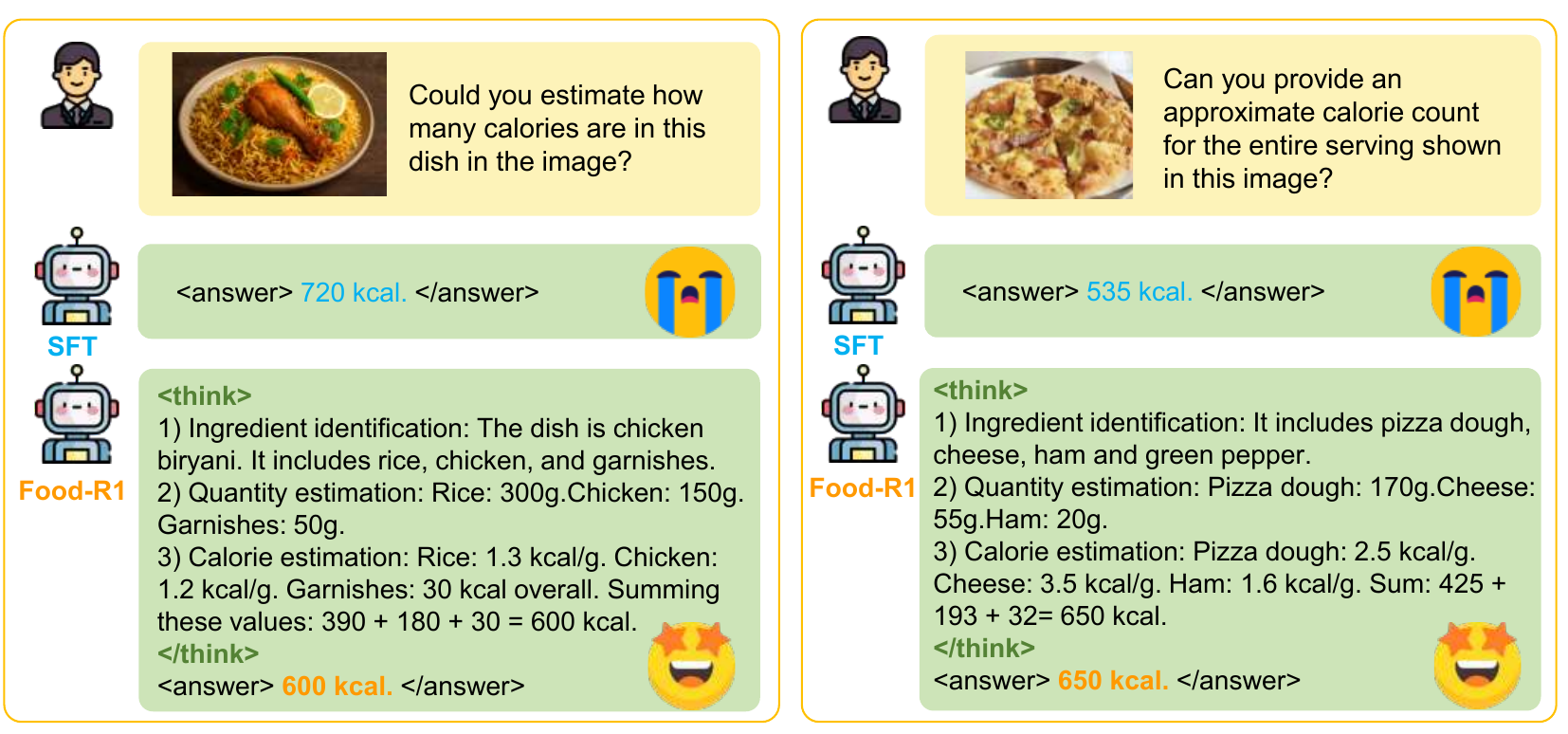}
  \caption{\textbf{Qualitative comparison before and after RFT.}
  Compared with the SFT-only model, Food-R1 after post-training RFT produces more stable and accurate calorie estimates through a step-by-step reasoning process.}
  \label{fig:calorie_think}
\end{figure}

Beyond model-side limitations, progress in food analysis is also constrained by data-related challenges. Existing datasets are often collected from heterogeneous sources, introducing non-food images and noisy annotations that require additional cleaning. Moreover, high-quality large-scale nutritional annotations remain scarce. 
Nutrition5k~\cite{thames2021nutrition5k} provides fine-grained labels but has limited category coverage, while CalData~\cite{yao2024caloraify}, built on Recipe1M+~\cite{marin2019recipe1mplus}, offers larger-scale supervision but relies on less reliable database-derived nutrition labels. These issues highlight the need for large-scale data with reliable nutritional information.

\begin{figure*}[t]
  \centering
  \includegraphics[width=0.95\textwidth]{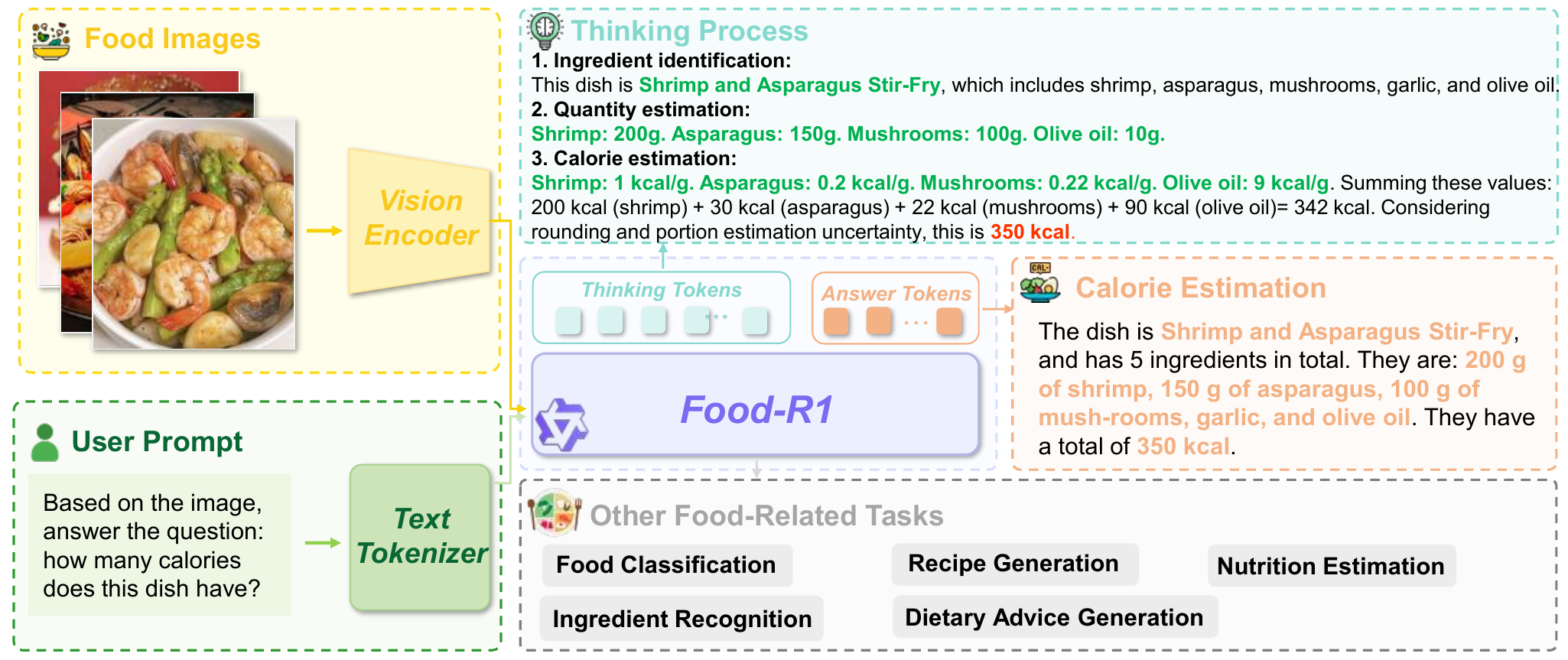}
  \caption{\textbf{Overview of Food-R1.}
  Given an image and text prompt, Food-R1 performs calorie estimation via CoT reasoning and outputs structured predictions, while also supporting food classification, ingredient recognition, recipe generation, dietary advice generation, and nutrition estimation.}
  \label{fig:overall_arch}
\end{figure*}

Motivated by these observations, we construct \textbf{CalorieBench-80K}, a large-scale benchmark built upon MM-Food-100K~\cite{dong2025mmfood100k}. Through systematic filtering, granularity alignment, and dietary advice annotation, CalorieBench-80K provides more reliable calorie-related supervision. To our knowledge, it is the first food image benchmark to incorporate Chain-of-Thought (CoT) annotations for calorie reasoning. 
We also propose \textbf{Food-R1}, a unified multi-task food VLM with various capabilities including calorie estimation, dietary advice generation, food classification, ingredient recognition, recipe generation, and nutrition estimation. Fig.~\ref{fig:overall_arch} summarizes our framework and highlights three key innovations: 
(1) \textbf{Unified Multi-task Food VLM}: to enhance overall capabilities and better support calorie estimation, we take CalorieBench-80K as the core and integrate representative datasets into a multi-task learning paradigm; 
(2) \textbf{Reasoning-oriented Distillation from Large Models}: to mitigate ambiguity in nutritional inference, we employ a strong teacher model to distill CoT rationales, making intermediate inference steps explicit; 
(3) \textbf{Food-based Reinforcement Learning}: to move beyond imitation learning, we apply reinforcement learning (RL) post-training using Group Relative Policy Optimization (GRPO), encouraging more reliable reasoning trajectories and improving overall performance.

We evaluate Food-R1 on CalorieBench-80K and other established benchmarks, where it achieves consistent improvements across food-related tasks and outperforms strong baselines. Our main contributions are as follows:
\begin{itemize}
    \item We introduce \textbf{CalorieBench-80K}, a large-scale benchmark with curated calorie labels and dietary advice annotations. To the best of our knowledge, it is the first food image benchmark to incorporate Chain-of-Thought annotations for calorie reasoning.
    \item We propose \textbf{Food-R1}, a unified multi-task food VLM. It integrates CoT-based reasoning distillation to make intermediate inference steps explicit, improving reasoning consistency and cross-task generalization.
    \item We further introduce a post-training reinforcement learning stage with Group Relative Policy Optimization to enhance the model’s reasoning capability and overall performance.
\end{itemize}

\section{Related Work}

\begin{figure*}[t]
  \centering
  \includegraphics[width=0.95\textwidth]{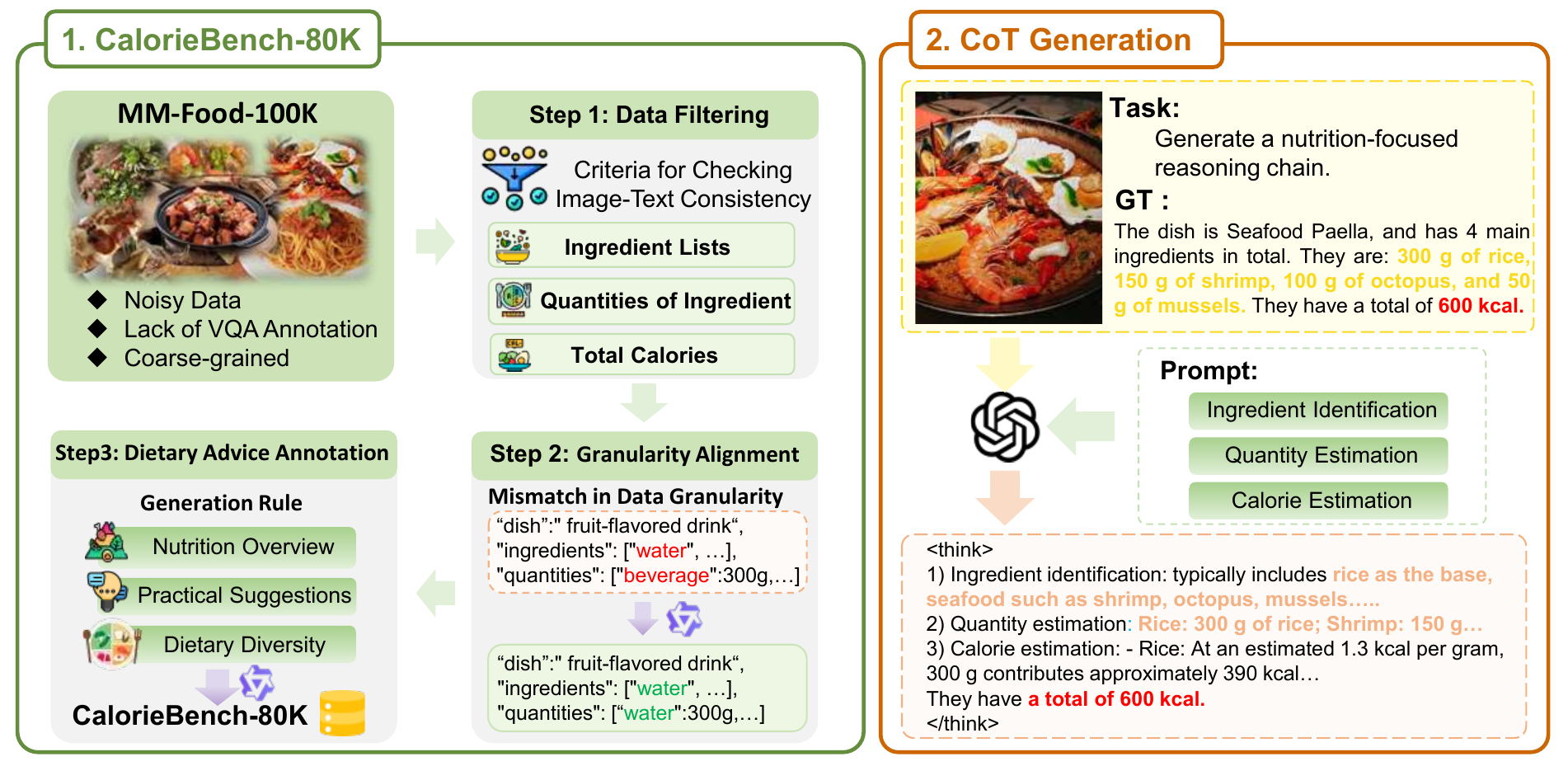}
  \caption{\textbf{CalorieBench-80K construction and CoT generation pipeline.}
  \textbf{Left:} Built upon MM-Food-100K, we apply data filtering, granularity alignment, and dietary advice annotation to obtain CalorieBench-80K. \textbf{Right:} We prompt ChatGPT to produce nutrition-focused CoT rationales for calorie estimation.}
  \label{fig:data_benchmark}
\end{figure*}

\subsection{Vision-Language Models for Food Analysis}
With improved multimodal understanding and reasoning, VLMs have significantly improved performance on food-related tasks. LLaVA-Chef~\cite{mohbat2024llavachef} employs fine-tuning and prompt engineering for recipe generation. Later studies~\cite{yao2024caloraify,liu2025retrievalaugmentedrecipe} adopt LoRA-based~\cite{hu2022lora} supervised fine-tuning for tasks such as nutrition estimation and recipe generation. FoodLMM~\cite{yin2024foodlmm} presents a versatile food assistant, and RoDE~\cite{jiao2024rode} introduces a multi-task learning framework for the food domain. However, most existing approaches remain centered on SFT. In contrast, as shown in Fig.~\ref{fig:sft_rft}, we adopt a two-stage training pipeline that consistently improves reasoning capabilities and performance across food-related tasks.

\subsection{Chain-of-Thought for Multimodal Reasoning}
Chain-of-Thought (CoT) reasoning encourages models to produce step-by-step rationales by exposing intermediate inference steps. It has been extended to multimodal tasks, including mathematical reasoning~\cite{zhang2024mavis}, robotic planning~\cite{mu2023embodiedgpt}, and autonomous driving~\cite{jiang2025alphadrive,li2025recogdrive}. Moreover, recent studies have explored structured, stage-wise reasoning traces for VLM training. For instance, LLaVA-CoT~\cite{xu2025llavacot} adopts structured CoT, while Corvid~\cite{jiang2025corvid} performs two-stage CoT-formatted fine-tuning; both yield notable gains. However, CoT-integrated VLM training for food analysis remains underexplored. In this work, we incorporate reasoning-oriented CoT distillation into training. Ablation results in Table~\ref{tab:mmfood_ablation} show that this component enhances overall performance.

\subsection{GRPO-based Reinforcement Learning for Reasoning}
Reinforcement learning (RL) is widely used in post-training to improve reasoning and overall model quality. DeepSeekMath~\cite{shao2024deepseekmath} introduced GRPO, which estimates advantages from within-group relative rewards and thus removes the need for an explicit value-function critic. This design improves training efficiency and stability. Recent R1-style works have extended GRPO-based training to multimodal tasks, including multimodal reasoning~\cite{huang2025visionr1}, chart comprehension~\cite{chen2025chartr1}, and autonomous driving~\cite{jiang2025alphadrive,li2025recogdrive}. Motivated by these advances, we adopt GRPO-based RL as a post-training stage for food-related tasks, achieving consistent gains in reasoning quality and task accuracy.

\section{Dataset Construction}

\subsection{Overview}
To obtain large-scale and reliable nutrition-related annotations, we construct CalorieBench\allowbreak-80K by systematically filtering and aligning samples from MM-Food-100K~\cite{dong2025mmfood100k} and then augmenting the data with dietary advice annotations, as illustrated in Fig.~\ref{fig:data_benchmark}. Step-wise human validation is conducted to assess the quality of filtering, granularity alignment, and dietary advice annotation.

\subsection{Step 1: Data Filtering}
Although MM-Food-100K provides a solid foundation, we observe issues such as image--text mismatches and noisy calorie annotations. To improve annotation reliability, we use GPT-4.1 to validate each sample against its food image from three aspects: (1) \textbf{Ingredient Consistency}, checking whether the listed ingredients are plausible and complete; (2) \textbf{Portion Plausibility}, checking whether the ingredient amounts match the visible portions; and (3) \textbf{Calorie Plausibility}, checking whether the total calories are reasonable. After filtering, approximately 80K high-confidence samples are retained. To verify filtering reliability, we manually review 500 sampled cases, including 300 retained and 200 filtered samples, following the same criteria, and observe a 93\% agreement between GPT-4.1's filtering decisions and human judgments.

\subsection{Step 2: Granularity Alignment}
We observe annotation granularity mismatches in some samples. For example, fine-grained ingredients (e.g., \textit{water}, \textit{sugar}) may be paired with only a coarse portion label (e.g., \textit{drink}: 300\,g), making it difficult to assign quantities to individual ingredients. To address this issue, we use Qwen2.5-VL-72B to map coarse portion quantities to the most plausible ingredients, producing aligned ingredient--quantity supervision. To verify alignment quality, we manually review 200 aligned samples and observe a 90\% agreement between model predictions and human judgments.

\subsection{Step 3: Dietary Advice Annotation}
To bridge calorie estimation and practical health guidance, we further augment the dataset with dietary advice annotations. We use Qwen2.5-VL-72B to generate structured advice covering three aspects: (1) an overall nutrition overview, (2) practical suggestions, and (3) encouraging dietary diversity. This provides supervision for actionable dietary advice generation. To ensure annotation quality, we manually reviewed 500 randomly sampled advice annotations. Most sampled annotations satisfy the required criteria, with no unsafe guidance observed.

\begin{figure*}[t]
  \centering
  \includegraphics[width=0.95\textwidth]{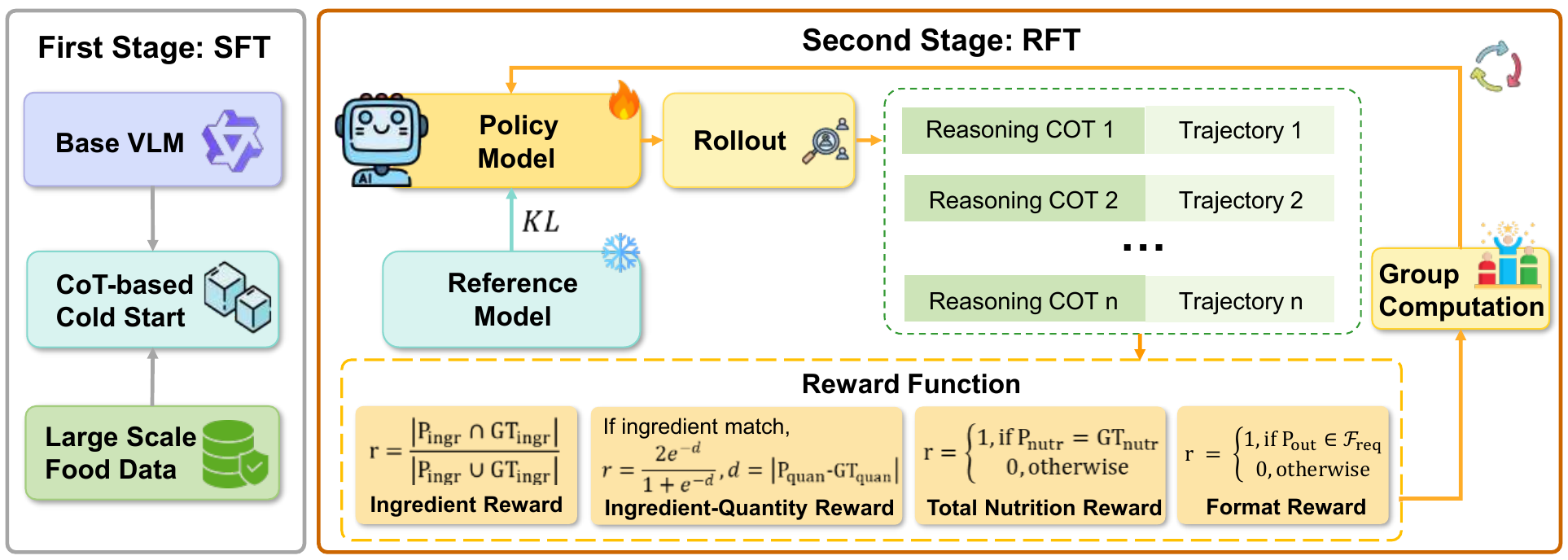}
  \caption{\textbf{Two-stage Training Paradigm.}
  Stage 1 (SFT): The base model is fine-tuned on mixed food datasets, incorporating CoT data for cold-start instruction tuning. Stage 2 (RFT): The SFT checkpoint serves as the reference model, while the policy model is optimized with GRPO using task-specific rewards.}
  \label{fig:sft_rft}
\end{figure*}

\section{Method}
\subsection{Overview} 
We aim to develop a general food VLM with reliable, nutrition-aware reasoning. We treat calorie estimation as the central task, since it requires multiple capabilities, including dish recognition, ingredient identification, and nutritional knowledge. To this end, we adopt a multi-task learning paradigm with CalorieBench-80K and auxiliary datasets, equipping the model with broad skills while strengthening calorie estimation. As shown in Fig.~\ref{fig:sft_rft}, we propose a two-stage training pipeline: Stage~1 performs CoT-based mixed SFT for cold-start instruction tuning, and Stage~2 applies GRPO-based RL for further performance improvement.

\subsection{Unified Multi-task Food VLM}
Food-R1 is a unified multi-task food VLM that maps a food image and an instruction into task-specific outputs in a shared autoregressive text generation space. Calorie estimation is treated as the central task, and complementary tasks, including food classification, ingredient recognition, recipe generation, nutrition estimation, and dietary advice generation, are integrated into a multi-task learning paradigm. By sharing visual semantics and task knowledge, this formulation enables cross-task transfer, improving the stability of calorie reasoning and overall generalization. All tasks follow an instruction-following format, and different tasks can be invoked by simply switching the instruction. The outputs range from structured numerical predictions to open-ended generation.

\subsection{Distillation from Large Models for Reasoning}
Supervising models only on final answers often fails to yield stable and interpretable reasoning. Therefore, as shown in Fig.~\ref{fig:data_benchmark}, we perform reasoning-oriented distillation by prompting GPT-4o to generate CoT annotations along three dimensions for a subset of CalorieBench-80K: (1) \textbf{Ingredient Identification}: recognizing the dish and main ingredients; (2) \textbf{Quantity Estimation}: assigning plausible amounts to each ingredient; (3) \textbf{Calorie Estimation}: inferring the energy density of each ingredient and aggregating them into the total calories.

The distilled CoT data are incorporated into SFT for cold-start instruction tuning. This encourages the model to decompose calorie estimation into interpretable subproblems and learn a more stable step-by-step solution trajectory, providing a stronger initialization for the subsequent RL stage.

\subsection{Food-based Reinforcement Learning}
Although SFT enables initial task decomposition, it remains insufficient for challenging food scenarios. We therefore apply GRPO-based RL post-training with task-specific rewards to stabilize long-chain reasoning and enhance performance.

\subsubsection{Group Relative Policy Optimization.}
For each training sample $(x,y)\sim\mathcal{D}$, we form a query $q$ and sample $G$ responses $\{o_i\}_{i=1}^{G}$ from $\pi_{\theta_{\text{old}}}(\cdot\mid q)$. Here $\pi_{\text{ref}}$ is the frozen SFT reference policy. We update $\pi_\theta$ by maximizing:
\begin{equation}
\small
\mathcal{J}_{\text{GRPO}}(\theta)=
\mathbb{E}_{(x,y)\sim\mathcal{D},\,\{o_i\}\sim \pi_{\theta_{\text{old}}}(\cdot\mid q)}
\bigl[
\frac{1}{G}\sum_{i=1}^{G}\mathcal{L}_i
-\beta D_{\mathrm{KL}}\!\left(\pi_\theta(\cdot\mid q) \,\|\, \pi_{\text{ref}}(\cdot\mid q)\right)
\bigr],
\end{equation}

\begin{equation}
\mathcal{L}_i
=
\min\!\left(
w_i A_i,\,
\mathrm{clip}(w_i, 1-\epsilon, 1+\epsilon)\,A_i
\right),
\end{equation}
where $w_i=\frac{\pi_\theta(o_i\mid q)}{\pi_{\theta_{\text{old}}}(o_i\mid q)}$ is the sequence-level likelihood ratio, $\epsilon$ and $\beta$ are hyperparameters, and $A_i$ is obtained by normalizing rewards within the group.

\subsubsection{Reward Design.}
We design four reward functions for ingredient recognition, and nutrition and calorie estimation.

\begin{itemize}  
  \item \textbf{Ingredient Match Reward.}
  This reward encourages accurate ingredient recognition, which is essential for calorie estimation. It is defined as the IoU between the predicted ingredient set $S_{\text{pred}}$ and the reference set $S_{\text{ref}}$.  

  \item \textbf{Ingredient--Quantity Match Reward.}
  This reward encourages plausible portion sizes for ingredients by matching $(\texttt{quantity}, \texttt{unit}, \texttt{ingredient})$ triplets between the prediction and the reference, thereby facilitating calorie estimation. For each matched triplet with quantity values \(\tilde{v}_{\text{pred}}\) and \(\tilde{v}_{\text{ref}}\), we define a smooth score:
  \begin{equation}
  r_{\text{qty}} =
  \frac{2 e^{-d}}{1 + e^{-d}},
  \quad
  d = \bigl|\tilde{v}_{\text{pred}} - \tilde{v}_{\text{ref}}\bigr|.
  \label{eq:reward-quantity}
  \end{equation}

  \item \textbf{Total Nutrition Match Reward.}
  Designed for calorie and nutrition estimation, this reward is computed as the fraction of correctly predicted required fields for the current dataset (e.g., calories only, or calories plus mass and macronutrients). With calories only, it reduces to a binary $0/1$ reward.

  \item \textbf{Format Reward.}
  To regularize the output format, we use a binary reward $r_{\text{fmt}}\in\{0,1\}$ that checks whether the output matches the required schema.
\end{itemize}

\section{Experiments}
\subsection{Experimental Settings}
\subsubsection{Summary of Datasets.}
Beyond CalorieBench-80K, we also incorporate representative datasets across food-related tasks, including food classification, recipe generation, and nutrition estimation. 

Food-101~\cite{bossard2014food101} and VireoFood-172~\cite{VireoFood172} are established benchmarks for food classification. VireoFood-172 and Recipe1M~\cite{salvador2017crossmodalrecipes} provide ingredient recognition supervision. For recipe generation, we use the image--recipe subset of Recipe1M, with paired recipes as supervision. Nutrition5k~\cite{thames2021nutrition5k} is used for nutrition estimation with RGB images only, and FoodDialogues~\cite{yin2024foodlmm} provides multi-turn conversational supervision. CalorieBench-80K follows a 75K/5K train/test split, and all auxiliary datasets use their official splits.

\subsubsection{Implementation Details.}
We use Qwen3-VL-8B as the base model and conduct SFT and GRPO-based RL post-training using the MS-SWIFT framework~\cite{zhao2025swift}. In the SFT stage, we train on all datasets in an answer-only instruction-following format and additionally include a 5K CoT subset from CalorieBench-80K to encourage interpretable intermediate reasoning. We perform full-parameter fine-tuning for 3 epochs with a learning rate of $4\times10^{-5}$ and a global batch size of 128 with 16 gradient accumulation steps.

For RL post-training, we initialize from the SFT checkpoint and run GRPO for 1 epoch with a global batch size of 128. We use a learning rate of $1\times10^{-6}$, sample 8 completions per prompt, and set the KL coefficient $\beta$ to 0.04. The rewards are equally weighted, with all coefficients set to 1. We exclude dietary advice, recipe generation, and multi-turn dialogue from GRPO, as their open-ended outputs make automatic reward evaluation unstable.

\begin{table}[t]
\caption{\textbf{Calorie estimation results on CalorieBench-80K.} \textbf{FoodLMM-s1} denotes FoodLMM trained on CalorieBench-80K with LoRA.}
\label{tab:calorie_results}
\centering
\small
\setlength{\tabcolsep}{4pt}
\renewcommand{\arraystretch}{1.05}
\begin{tabular}{@{}lrrr@{}}
\toprule
\textbf{Model} & \textbf{MAE} $\downarrow$ & \textbf{RMSE} $\downarrow$ & \textbf{$R^2$} $\uparrow$ \\
\midrule
Gemini-2.5-flash & 301.87   &  576.65   &  -5.77      \\
InternVL3-8B     & 218.70   &  412.16   &  -1.68      \\
LLaVA-7B         & 200.79   &  304.21   &  -0.48      \\
Qwen2.5-VL-72B   & 120.95   &  227.98   &   0.18      \\
Qwen3-VL-8B      & 195.18   &  370.04   &  -1.13      \\
GPT-4o-mini      &  88.40   &  167.00   &   0.57      \\
FoodLMM~\cite{yin2024foodlmm} & 170.35  & 274.22  &  -0.19  \\
FoodLMM-s1       & 57.99    & 134.46    &   0.71  \\
\textbf{Food-R1} & \textbf{42.56} & \textbf{112.89} & \textbf{0.81} \\
\bottomrule
\end{tabular}
\end{table}

\begin{table}[t]
\caption{\textbf{Dietary advice generation results on CalorieBench-80K.}
We report ROUGE scores as R (i.e., R-1/R-2/R-L/R-L$_{\text{sum}}$), BLEU, SacreBLEU, and BERTScore as BERT (i.e., BERT-P/BERT-R/BERT-F1).}
\label{tab:diet_advice_results}
\centering
\small
\setlength{\tabcolsep}{2.5pt}
\renewcommand{\arraystretch}{1.08}
\resizebox{\columnwidth}{!}{%
\begin{tabular}{@{}lccccccccc@{}}
\toprule
Model & R-1 $\uparrow$ & R-2 $\uparrow$ & R-L $\uparrow$ & R-L$_{\text{sum}}$ $\uparrow$ &
BLEU $\uparrow$ & SacreBLEU $\uparrow$ &
BERT-P $\uparrow$ & BERT-R $\uparrow$ & BERT-F1 $\uparrow$ \\
\midrule
Gemini-2.5-flash      & 0.42 & 0.13 & 0.24 & 0.24 & 0.06 & 0.10 & 0.86 & 0.85 & 0.85 \\
InternVL3-8B          & 0.55 & 0.21 & 0.33 & 0.33 & 0.11 & 0.17 & 0.90 & 0.91 & 0.91 \\
LLaVA-7B              & 0.52 & 0.19 & 0.30 & 0.30 & 0.06 & 0.09 & 0.89 & 0.89 & 0.89 \\
Qwen2.5-VL-72B        & 0.45 & 0.17 & 0.29 & 0.29 & 0.07 & 0.07 & 0.89 & 0.88 & 0.88 \\
Qwen3-VL-8B           & 0.47 & 0.13 & 0.25 & 0.25 & 0.04 & 0.05 & 0.87 & 0.88 & 0.88 \\
GPT-4o-mini           & 0.55 & 0.21 & 0.33 & 0.33 & 0.15 & 0.21 & 0.91 & 0.91 & 0.91 \\
FoodLMM               & 0.23 & 0.04 & 0.12 & 0.12 & 0.01 & 0.02 & 0.84 & 0.82 & 0.83 \\
FoodLMM-s1            & 0.63 & 0.34 & 0.44 & 0.44 & 0.22 & 0.27 & 0.92 & 0.91 & 0.91 \\
\textbf{Food-R1} & \textbf{0.65} & \textbf{0.39} & \textbf{0.49} & \textbf{0.49} & \textbf{0.32} & \textbf{0.40} & \textbf{0.93} & \textbf{0.93} & \textbf{0.93} \\
\bottomrule
\end{tabular}%
}
\end{table}

\begin{table}[t]
\caption{\textbf{Results on food classification and ingredient recognition.}
Food-R1 achieves the best performance on both Food-101 classification and VireoFood-172 ingredient recognition.}
\label{tab:food101_vireo172}
\centering
\small
\setlength{\tabcolsep}{4pt}
\renewcommand{\arraystretch}{1.08}
\begin{tabular}{@{}lc@{\hspace{18pt}}lcc@{}}
\toprule
\multicolumn{2}{c}{\textbf{Food-101 Classification}} 
& \multicolumn{3}{c}{\textbf{VireoFood-172 Ingredient Recognition}} \\
\cmidrule(lr){1-2} \cmidrule(lr){3-5}
Method & Top-1 Acc. (\%) $\uparrow$
& Method & IoU $\uparrow$ & F1 $\uparrow$ \\
\midrule
PRENet~\cite{min2023largescalevisualfood_tpami} 
& 91.13
& CACLNet~\cite{luo2023ingredientpredictioncln}
& -- & 65.71 \\

FoodLMM 
& 93.93
& FoodLMM
& 56.94 & 68.97 \\

RoDE~\cite{jiao2024rode}
& 94.02
& RoDE~\cite{jiao2024rode}
& 65.95 & 73.54 \\

\textbf{Food-R1}
& \textbf{95.21}
& \textbf{Food-R1}
& \textbf{78.50} & \textbf{84.61} \\
\bottomrule
\end{tabular}
\end{table}

\begin{table}[t]
\caption{\textbf{Recipe generation results on Recipe1M.} Food-R1 achieves the highest SacreBLEU, while LLaVA-FT performs best on ROUGE-L.}
\label{tab:recipe1m_gen}
\centering
\small
\setlength{\tabcolsep}{4pt}
\renewcommand{\arraystretch}{1.08}
\begin{tabular}{@{}lcc@{}}
\toprule
Method & SacreBLEU $\uparrow$ & ROUGE-L $\uparrow$ \\
\midrule
FIRE~\cite{chhikara2024fire}   & 6.02      & 21.29          \\
FoodLMM                        & 6.24      & 36.96          \\
LLaVA-FT~\cite{liu2025retrievalaugmentedrecipe} & 5.88  & \textbf{38.18} \\
\textbf{Food-R1}               & \textbf{7.76}  & 28.73          \\
\bottomrule
\end{tabular}
\end{table}

\begin{table}[t]
\caption{\textbf{Nutrition estimation results on Nutrition5k.} The Average column represents the mean MAE and pMAE across all five tasks.}
\label{tab:nutrition5k_total}
\centering
\small
\setlength{\tabcolsep}{2pt}
\renewcommand{\arraystretch}{1.08}

\resizebox{0.98\columnwidth}{!}{%
\begin{tabular}{@{\hspace{3pt}}lcccccc@{\hspace{3pt}}}
\toprule
Method & Calorie MAE $\downarrow$ & Mass MAE $\downarrow$ & Fat MAE $\downarrow$ & Carb MAE $\downarrow$ & Protein MAE $\downarrow$ & Average $\downarrow$ \\
\midrule
RoDE  & -- / 52.4\%  & -- / 38.4\%   & -- / 67.1\% & -- / 47.8\%  & -- / 53.9\%  & -- / 51.9\% \\
FoodLMM & 67.3 / 26.6\% & \textbf{39.7} / 20.7\% & 5.4 / 39.7\% & 5.9 / 31.1\% & 4.1 / 25.8\% & 24.5 / 28.8\% \\
\textbf{Food-R1} & \textbf{27.6 / 14.4\%} & 44.3 / \textbf{18.0\%} & \textbf{3.2 / 26.4\%} & \textbf{4.4 / 22.9\%} & \textbf{4.0 / 23.9\%} & \textbf{16.7 / 21.1\%} \\
\bottomrule
\end{tabular}%
}
\end{table}

\begin{table}[t]
\caption{\textbf{Referring nutrition estimation results on Nutrition5k.} Food-R1 achieves comparable or better performance across referenced-ingredient settings.}
\label{tab:nutrition5k_ref}
\centering
\small
\setlength{\tabcolsep}{4pt}
\renewcommand{\arraystretch}{1.08}

\resizebox{0.98\columnwidth}{!}{%
\begin{tabular}{@{\hspace{3pt}}clcccccc@{\hspace{3pt}}}
\toprule
Refer ingredient & Method
& Calorie MAE $\downarrow$
& Mass MAE $\downarrow$
& Fat MAE $\downarrow$
& Carb MAE $\downarrow$
& Protein MAE $\downarrow$
& Average $\downarrow$ \\
\midrule
\multirow{2}{*}{refer@1st}
& FoodLMM          & \textbf{25.2} / 34.7\% & 47.3 / 37.1\%          & 3.7 / 46.1\%          & 3.7 / 34.1\%      & 2.5 / 22.4\% & 16.5 / 34.9\% \\
& \textbf{Food-R1} & 28.2 / \textbf{21.4\%}          & \textbf{22.0 / 21.0\%} & \textbf{1.3 / 22.9\%} & \textbf{2.5 / 25.6\%} & \textbf{2.4 / 21.9\%}    & \textbf{11.3 / 22.6\%} \\
\midrule
\multirow{2}{*}{refer@2nd}
& FoodLMM   & \textbf{21.0 / 27.4\%} & \textbf{16.9} / 39.5\% 
& 1.1 / 29.0\%          & \textbf{2.1 / 30.5\%} & \textbf{1.4 / 27.4\%} & \textbf{8.5 / 30.8\%} \\
& \textbf{Food-R1} & 24.4 / 31.2\%    & 18.3 / \textbf{29.7\%}     & \textbf{1.0 / 27.5\%} & 2.5 / 34.3\%  & 1.6 / 32.6\% & 9.6 / 31.1\% \\
\midrule
\multirow{2}{*}{refer@3rd}
& FoodLMM          & 16.9 / 39.5\%          & \textbf{14.2} / 37.5\% & 0.9 / 39.9\%     & 1.5 / 38.1\% & 1.2 / 48.2\%           & 6.9 / 40.6\% \\
& \textbf{Food-R1} & \textbf{13.3 / 32.2\%} & 12.4 / \textbf{33.2\%}      & \textbf{0.6 / 28.5\%} & \textbf{1.4 / 36.5\%}      & \textbf{0.7 / 32.1\%} & \textbf{5.7 / 32.5\%} \\
\bottomrule
\end{tabular}%
}
\end{table}

\begin{figure}[!t]
  \centering
  \includegraphics[width=0.90\textwidth]{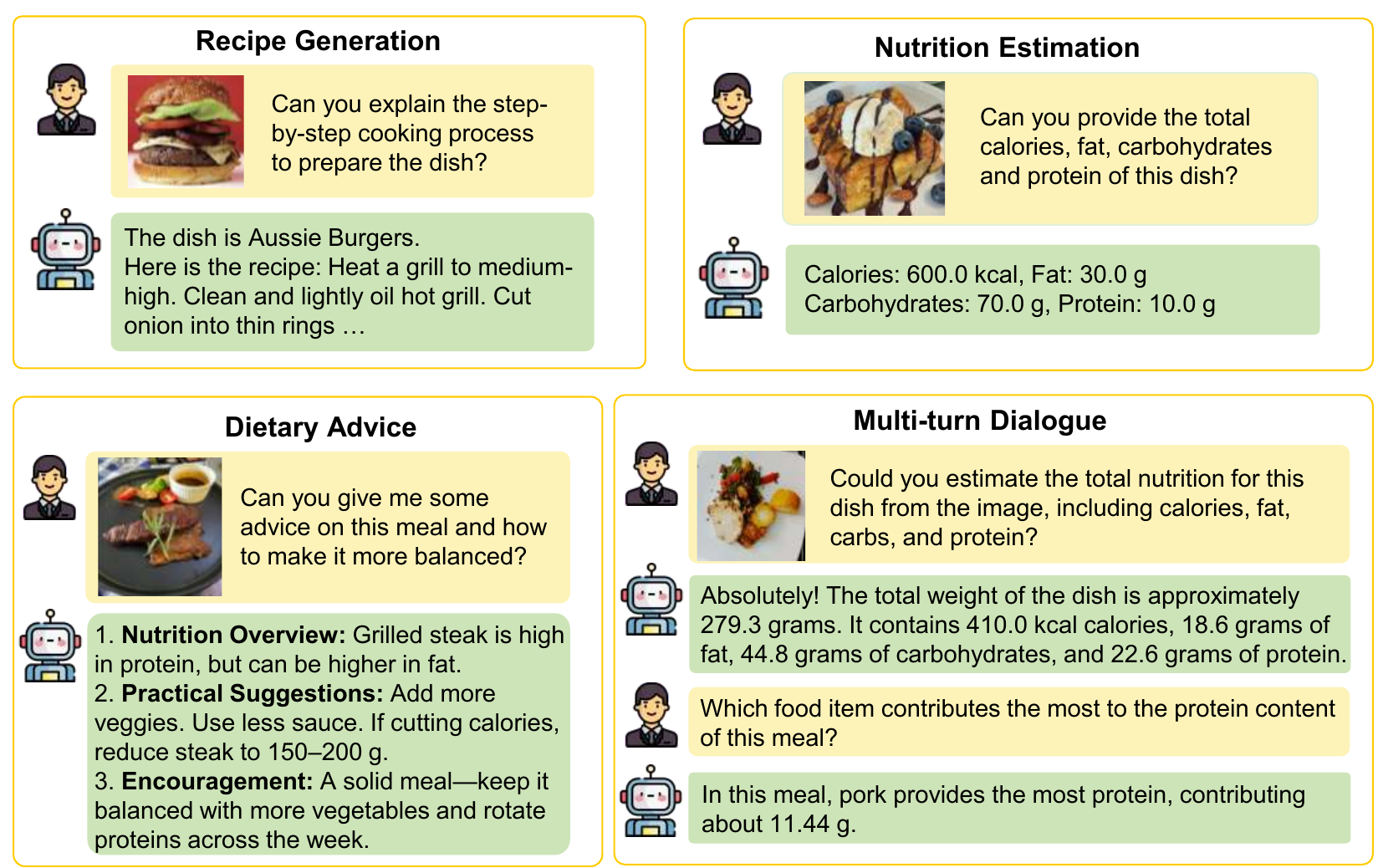}
  \caption{\textbf{Representative examples of Food-R1.}
  The examples cover recipe generation, nutrition estimation, dietary advice, and multi-turn dialogue.}
  \label{fig:foodlmm_exa}

  \vspace{0.4em}

  \captionof{table}{\textbf{Ablation study} of calorie estimation on CalorieBench-80K with progressively added training strategies. The Multi-task Learning paradigm refers to whether training includes auxiliary food datasets apart from CalorieBench-80K.}
  \label{tab:mmfood_ablation}
  \small
  \setlength{\tabcolsep}{3pt}
  \renewcommand{\arraystretch}{1.08}
  \begin{tabular*}{0.95\textwidth}{@{\extracolsep{\fill}}ccccccc@{}}
  \toprule
  \shortstack[c]{Supervised\\Fine-Tuning} &
  \shortstack[c]{Multi-task\\Learning} &
  \shortstack[c]{Chain-of-\\Thought} &
  \shortstack[c]{Reinforcement\\Learning} &
  MAE $\downarrow$ & RMSE $\downarrow$ & $R^2$ $\uparrow$ \\
  \midrule
  \cmark & \xmark & \xmark & \xmark & 51.62 & 121.56 & 0.77 \\
  \cmark & \cmark & \xmark & \xmark & 49.28 & 118.23 & 0.78 \\
  \cmark & \cmark & \cmark & \xmark & 45.28 & 113.44 & 0.80 \\
  \cmark & \cmark & \cmark & \cmark & \textbf{42.56} & \textbf{112.89} & \textbf{0.81} \\
  \bottomrule
  \end{tabular*}

  \vspace{-0.5em}
\end{figure}

\subsection{Evaluation Metrics}
We adopt task-specific metrics for each food-related task. For \textbf{calorie estimation}, we report MAE, RMSE, and $R^2$, following ~\cite{dong2025mmfood100k}. For \textbf{dietary advice generation}, we use ROUGE-1/2/L/Lsum~\cite{lin2004rouge}, BLEU~\cite{papineni-etal-2002-bleu}, SacreBLEU~\cite{post2018sacrebleu}, and BERTScore~\cite{zhang2020bertscore}. For \textbf{food classification}, we report top-1 accuracy. For \textbf{ingredient recognition}, we use Intersection over Union (IoU) and F1~\cite{manning2008ir}. For \textbf{recipe generation}, we measure the quality of generated recipes using SacreBLEU and ROUGE-L, following~\cite{yin2024foodlmm}. For \textbf{nutrition estimation}, we report MAE and percentage MAE (pMAE) for calories, mass, fat, carbohydrates, and protein, where pMAE is the MAE divided by the mean ground-truth value of each field. Calories are measured in kcal, and other fields in grams.

\subsection{Main Results}
\subsubsection{Calorie Estimation and Dietary Advice Generation Results.}
Table~\ref{tab:calorie_results} reports calorie estimation results on CalorieBench-80K. We compare Food-R1 with several zero-shot general-purpose VLMs. For domain-specific comparison, we report FoodLMM~\cite{yin2024foodlmm} under zero-shot inference and its LoRA fine-tuned variant \textbf{FoodLMM-s1}, trained on the CalorieBench-80K training set with LoRA ($r{=}8$, $\alpha{=}16$, dropout $0.05$). As shown, Food-R1 achieves the best overall performance, outperforming both the domain-specific baseline FoodLMM-s1 and the teacher VLM Qwen2.5-VL-72B used in data construction, demonstrating the effectiveness of our training pipeline.

Table~\ref{tab:diet_advice_results} reports dietary advice generation results on CalorieBench-80K. Food-R1 obtains the highest scores across most metrics, surpassing Qwen2.5-VL-72B and FoodLMM-s1, indicating stronger alignment with the reference advice.

\subsubsection{Food Classification and Ingredient Recognition Results.}
As shown in Table~\ref{tab:food101_vireo172}, Food-R1 achieves 95.21\% Top-1 accuracy on Food-101 and obtains an IoU of 78.50 and an F1 of 84.61 on VireoFood-172 ingredient recognition. These results show that Food-R1 outperforms existing methods on both food classification and fine-grained ingredient recognition, achieving state-of-the-art performance and validating the effectiveness of our multi-task training strategy.

\subsubsection{Recipe Generation Results.}
Table~\ref{tab:recipe1m_gen} reports recipe generation results on Recipe1M. Food-R1 achieves the highest SacreBLEU score, indicating better matching with reference recipes in local phrases and common cooking expressions. However, its ROUGE-L score is lower than that of LLaVA-FT, possibly because Food-R1 produces shorter recipes, while Recipe1M references often include more specific ingredient quantities and longer steps. Since ROUGE-L is more sensitive to long subsequence overlap, text length, and step order, the two metrics capture different aspects of generation quality.

\subsubsection{Nutrition Estimation Results.}
Table~\ref{tab:nutrition5k_total} reports results on Nutrition5k for estimating total calories, mass, and macronutrients from food images. Overall, Food-R1 achieves the best \textit{Average} performance, reducing the error over FoodLMM by 7.8 MAE and 7.7 pMAE points. Here, \textit{Average} denotes the mean error across the five targets.

We evaluate \emph{referring nutrition estimation}~\cite{yin2024foodlmm}, where the model predicts nutrition for a specified ingredient. Following FoodLMM, we use the top-3 ingredients by mass as references (refer@1st/\allowbreak2nd/\allowbreak3rd). As shown in Table~\ref{tab:nutrition5k_ref}, Food-R1 outperforms FoodLMM on refer@1st and refer@3rd, and achieves comparable results on refer@2nd, showing strong ingredient-level nutrition estimation.

\subsection{Ablation Study}
We conduct a stepwise ablation study on CalorieBench-80K, with results shown in Table~\ref{tab:mmfood_ablation}. Multi-task learning brings consistent improvements over the SFT baseline, indicating that auxiliary food datasets provide complementary supervision. Adding CoT supervision further reduces MAE from 49.28 to 45.28, suggesting that explicit reasoning helps calorie estimation. Finally, GRPO-based reinforcement learning further improves performance, leading Food-R1 to achieve the best results, with an MAE of 42.56, RMSE of 112.89, and $R^2$ of 0.81. These results demonstrate the effectiveness of our training strategy.
\section{Conclusion}

We introduce CalorieBench-80K, a large-scale benchmark with reliable calorie labels and dietary advice annotations. We further propose Food-R1, a unified food VLM trained in a multi-task learning paradigm with RL post-training. By leveraging CalorieBench-80K and auxiliary datasets, Food-R1 improves generalization across food-related tasks and outperforms strong baselines, validating the effectiveness of our approach.

\bibliographystyle{splncs04}
\bibliography{main}

\clearpage
\appendix
\FloatBarrier

% ---------- Appendix numbering ----------
\setcounter{section}{0}
\setcounter{subsection}{0}
\setcounter{subsubsection}{0}
\setcounter{figure}{0}
\setcounter{table}{0}

% Sections: A, B, C / A.1, A.2, ...
\renewcommand{\thesection}{\Alph{section}}
\renewcommand{\thesubsection}{\thesection.\arabic{subsection}}
\renewcommand{\thesubsubsection}{\thesubsection.\arabic{subsubsection}}

% Keep supplementary-style figure/table numbering.
\renewcommand{\thefigure}{S\arabic{figure}}
\renewcommand{\thetable}{S\arabic{table}}

\begin{center}
    {\Large\bfseries Appendix}
\end{center}

\vspace{0.8em}

\noindent\textbf{Overview.}
The appendix is organized as follows.
\textbf{Appendix~\ref{sec:supp-s1}} presents the construction details and human validation of CalorieBench-80K.
\textbf{Appendix~\ref{sec:supp-s2}} introduces the datasets used in training, the instruction templates for multi-task learning, and the subset constructed from CalorieBench\allowbreak-80K with Chain-of-Thought (CoT) annotations.
\textbf{Appendix~\ref{sec:supp-s3}} provides experimental details, including training hyperparameters, baseline configurations, and other evaluation settings.

\vspace{1em}

\begin{figure}[t]
  \centering
  \includegraphics[width=\textwidth]{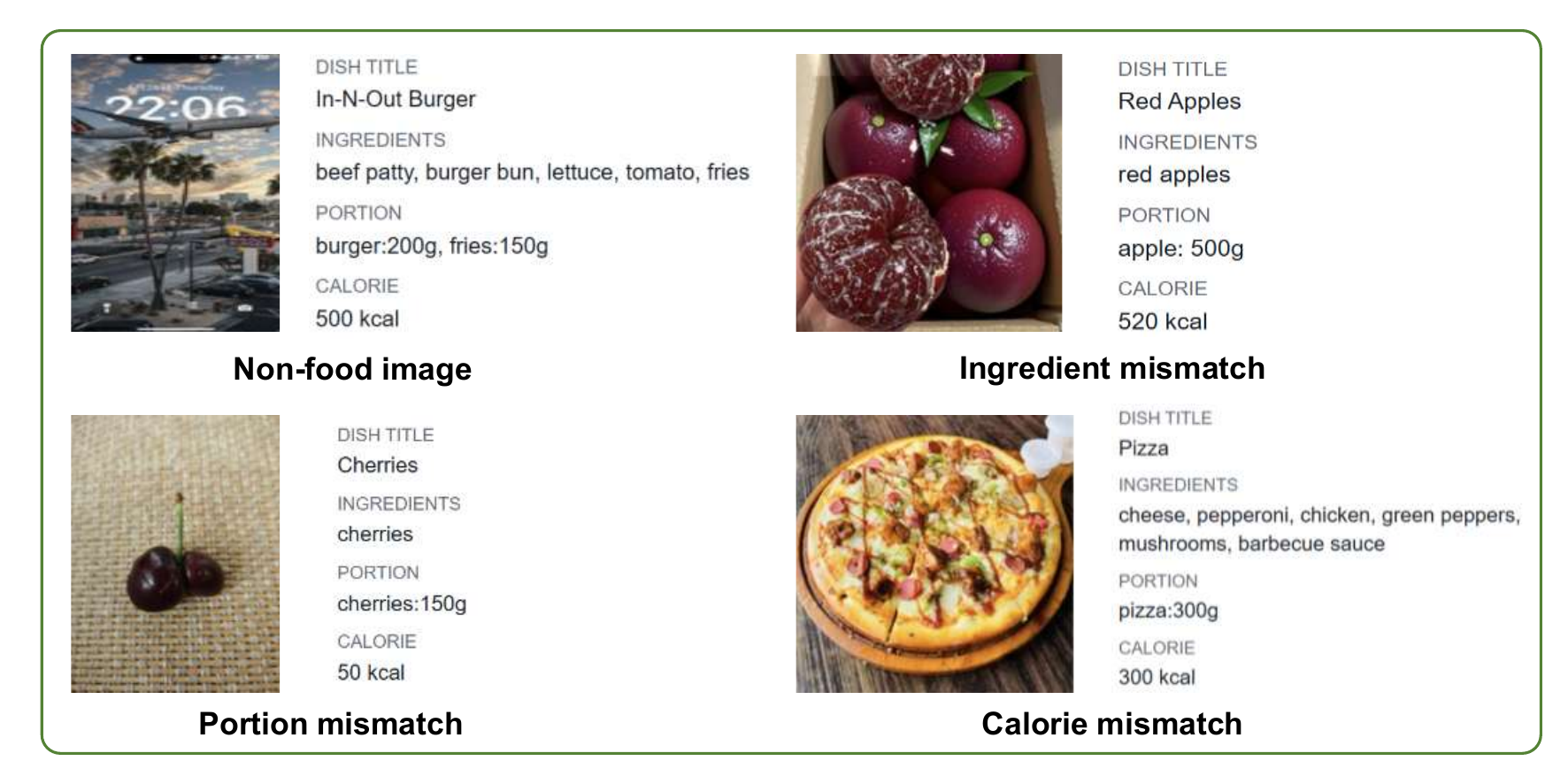}
  \caption{\textbf{Representative noisy samples removed by the filtering pipeline.}
  These examples show typical raw-data errors, including non-food images, ingredient mismatches, portion errors, and calorie inconsistencies.}
  \label{fig:supp-12}
\end{figure}

\begin{figure}[t]
  \centering
  \includegraphics[width=\textwidth]{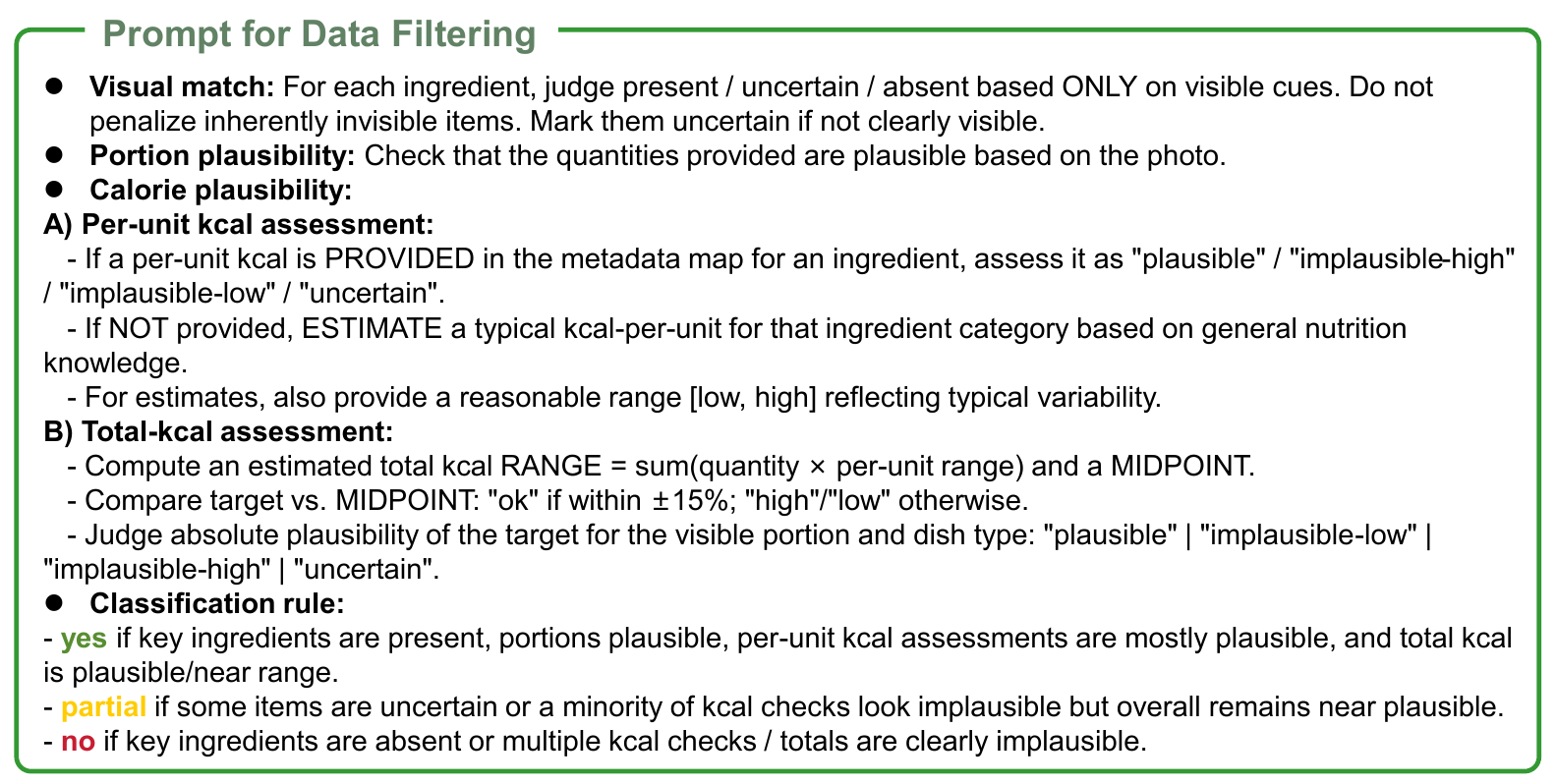}
  \caption{Prompt template for filtering MM-Food-100K samples.}
  \label{fig:supp-1}
\end{figure}

\begin{figure}[t]
  \centering
  \includegraphics[width=\textwidth]{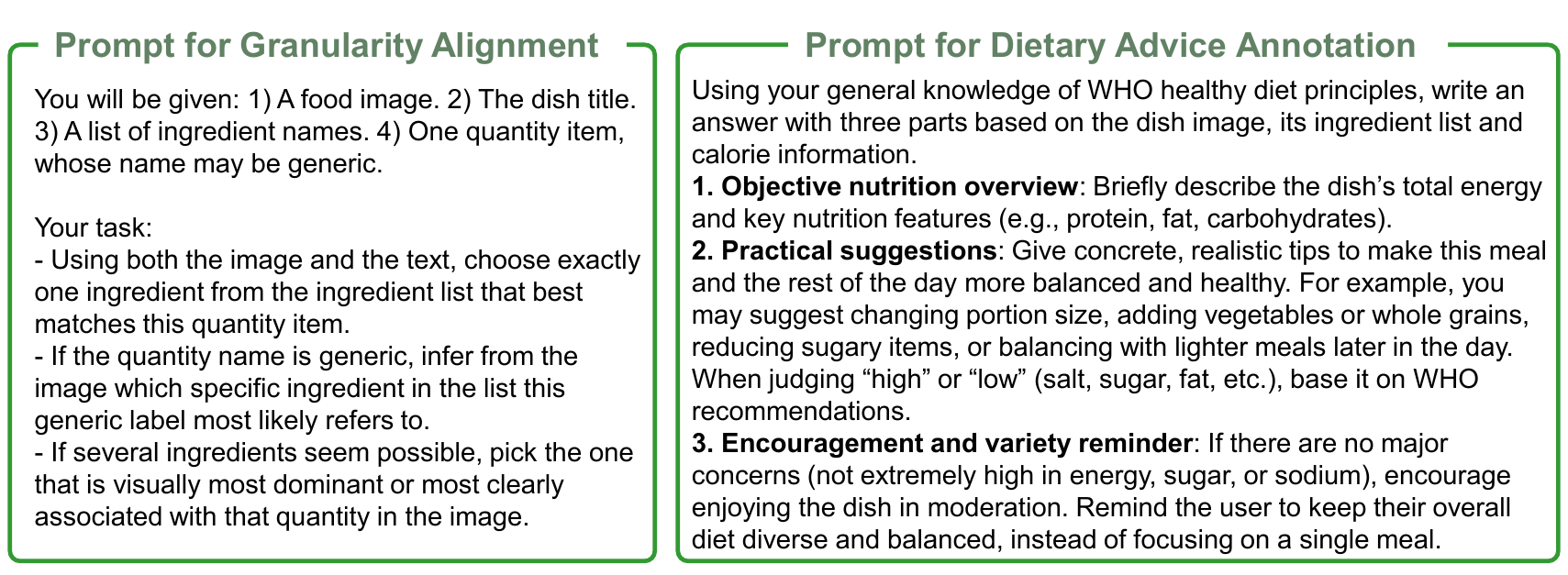}
  \caption{Prompt templates used for granularity alignment (left) and dietary advice annotation (right).}
  \label{fig:supp-2}
\end{figure}

\section{CalorieBench-80K Construction Details}\label{sec:supp-s1}
\subsection{Construction Pipeline}
As illustrated in Fig.~\ref{fig:supp-1}, we use GPT-4.1 to filter the MM-Food-100K samples~\cite{dong2025mmfood100k}. GPT-4.1 provides an overall plausibility assessment together with a confidence score. Under this criterion, we retain only samples labeled as ``yes'' or ``partial'' with a confidence score greater than 0.8, yielding approximately 80,000 high-confidence samples for subsequent training and evaluation. Fig.~\ref{fig:supp-12} shows representative noisy samples discarded by our filtering pipeline. The full prompts used in the construction pipeline are provided in Fig.~\ref{fig:supp-1} and Fig.~\ref{fig:supp-2}.

\subsection{Human Validation}
To verify the reliability of the CalorieBench-80K construction pipeline, we conduct human validation on three key steps: data filtering, granularity alignment, and dietary advice annotation.

For Step 1, data filtering, we randomly sample 500 instances for manual inspection, including 300 retained samples and 200 filtered samples. The manual validation follows the same criteria as the GPT-4.1-based filtering stage. The results show that the decision accuracy is 95.0\% for retained samples and 90.0\% for filtered samples, leading to an overall correctness of 93.0\%. This indicates that the filtering process can effectively remove many clearly implausible samples.

For Step 2, granularity alignment, we randomly sample 200 instances after granularity alignment and check whether each coarse portion is correctly mapped to an appropriate ingredient. The alignment accuracy is 90.0\%, suggesting that this step can generally establish reasonable correspondences between portion descriptions and specific ingredients.

For Step 3, dietary advice annotation, we randomly sample 500 instances with dietary advice and check whether they satisfy the annotation requirements. We further evaluate safety by examining whether the generated advice contains inappropriate content, such as medical diagnoses, disease risk assessments, or extreme dietary suggestions. All sampled instances pass the safety check, achieving a safety pass rate of 100.0\%. This suggests that the generated dietary advice annotations satisfy the predefined safety criteria.
\section{Datasets and Prompt Design Details}
\label{sec:supp-s2}
\subsection{Dataset Overview}
We use the following six datasets for training:

\begin{figure}[t]
  \centering
  \includegraphics[width=\textwidth]{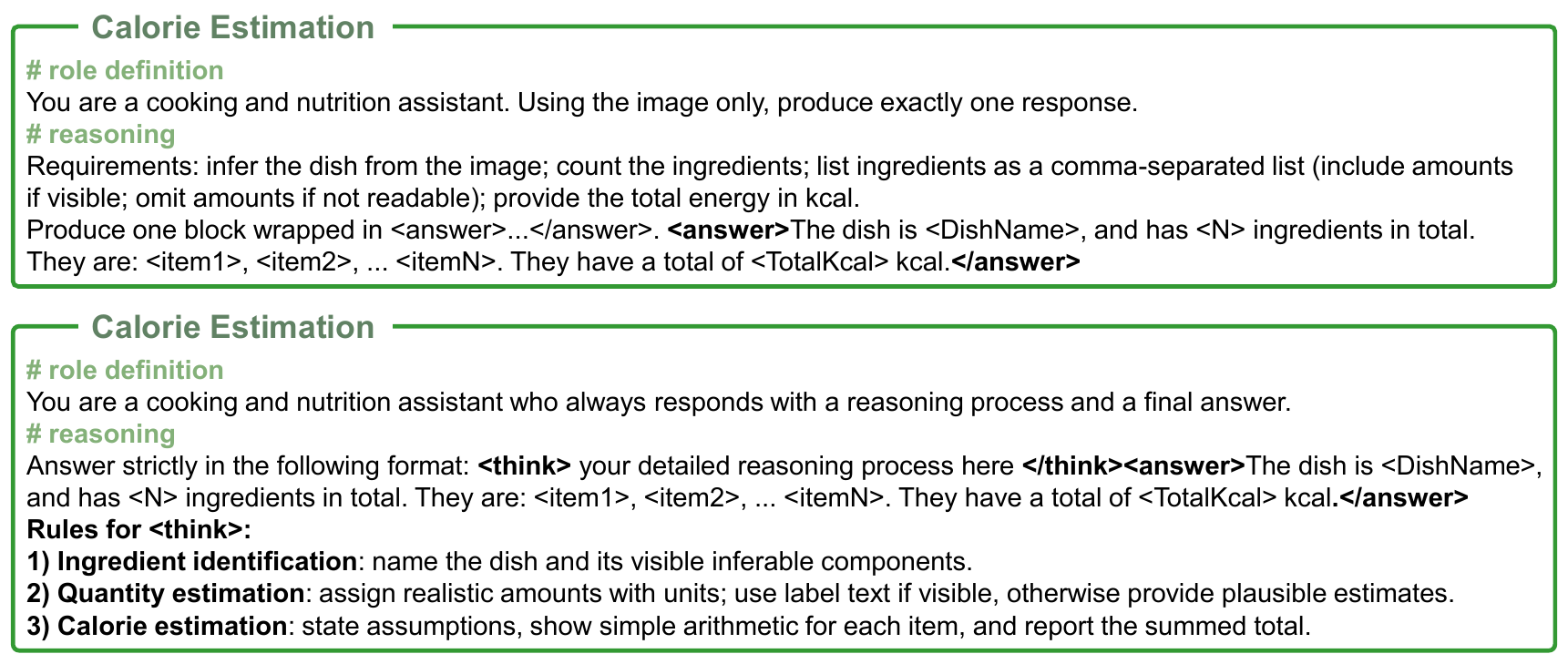}
  \caption{Prompt templates for calorie estimation in the answer-only format and the format with CoT.}
  \label{fig:supp-3}
\end{figure}

\begin{figure}[t]
  \centering
  \includegraphics[width=\textwidth]{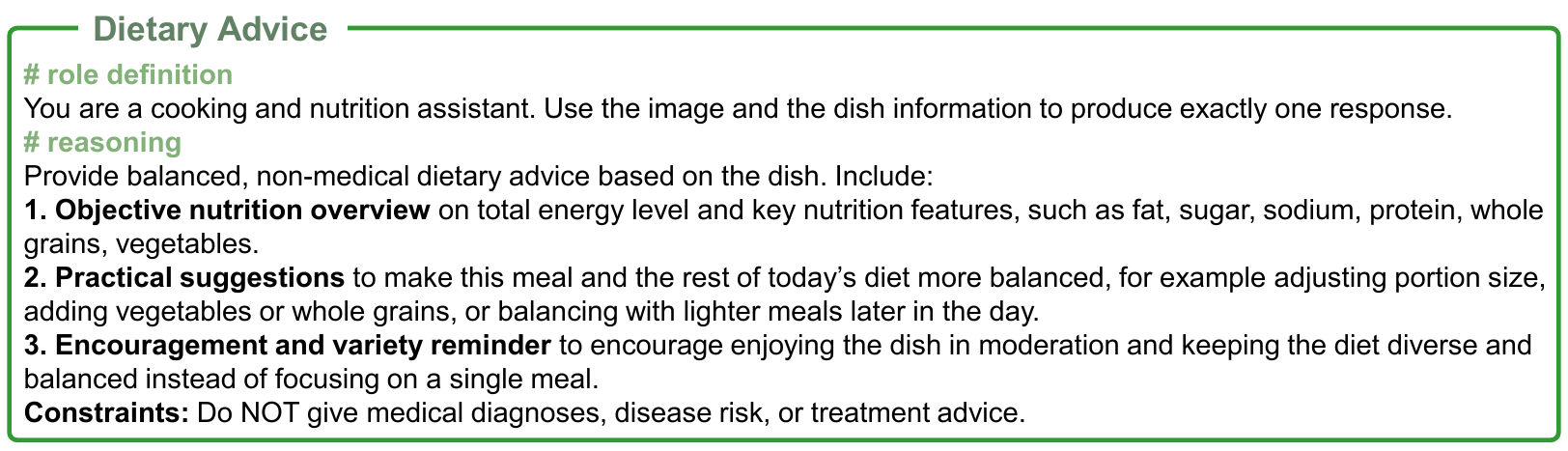}
  \caption{Prompt template for dietary advice generation.}
  \label{fig:supp-4}
\end{figure}

\begin{figure}[t]
  \centering
  \includegraphics[width=\textwidth]{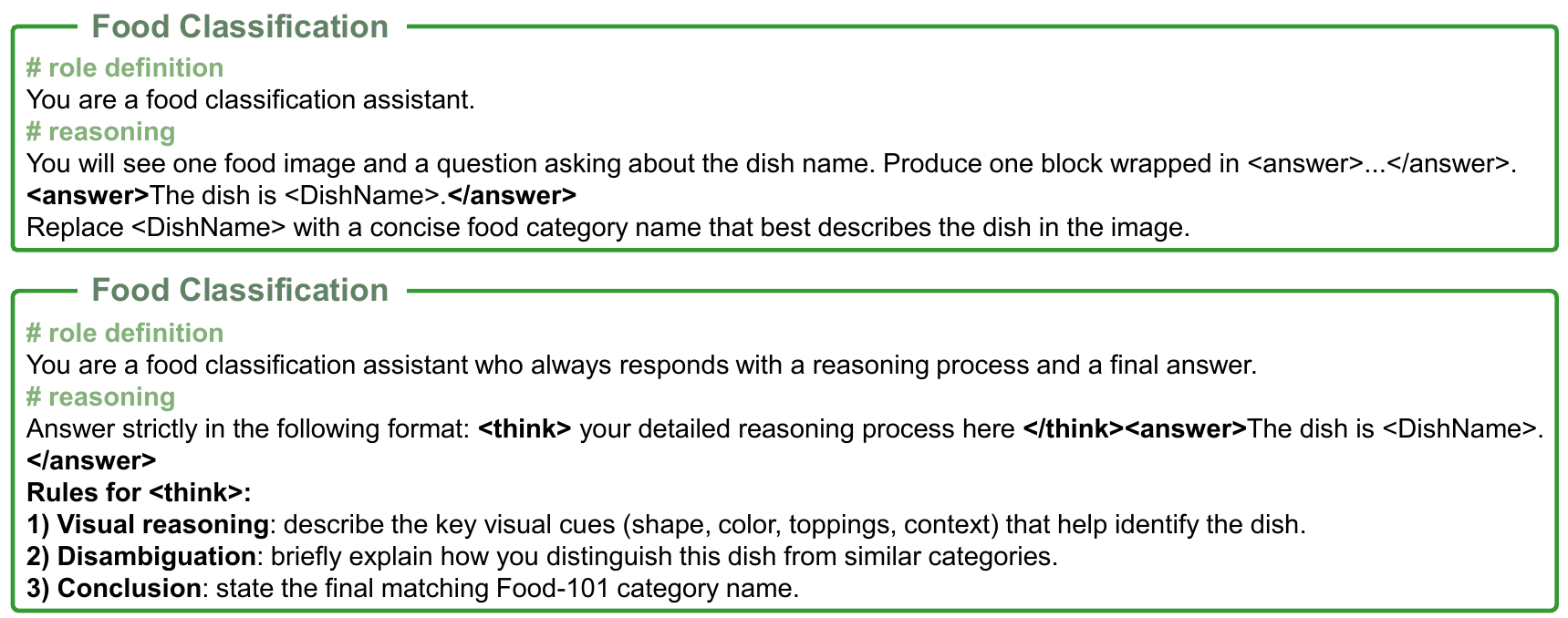}
  \caption{Prompt templates for food classification in the answer-only format and the format with CoT.}
  \label{fig:supp-5}
\end{figure}

\begin{figure}[t]
  \centering
  \includegraphics[width=\textwidth]{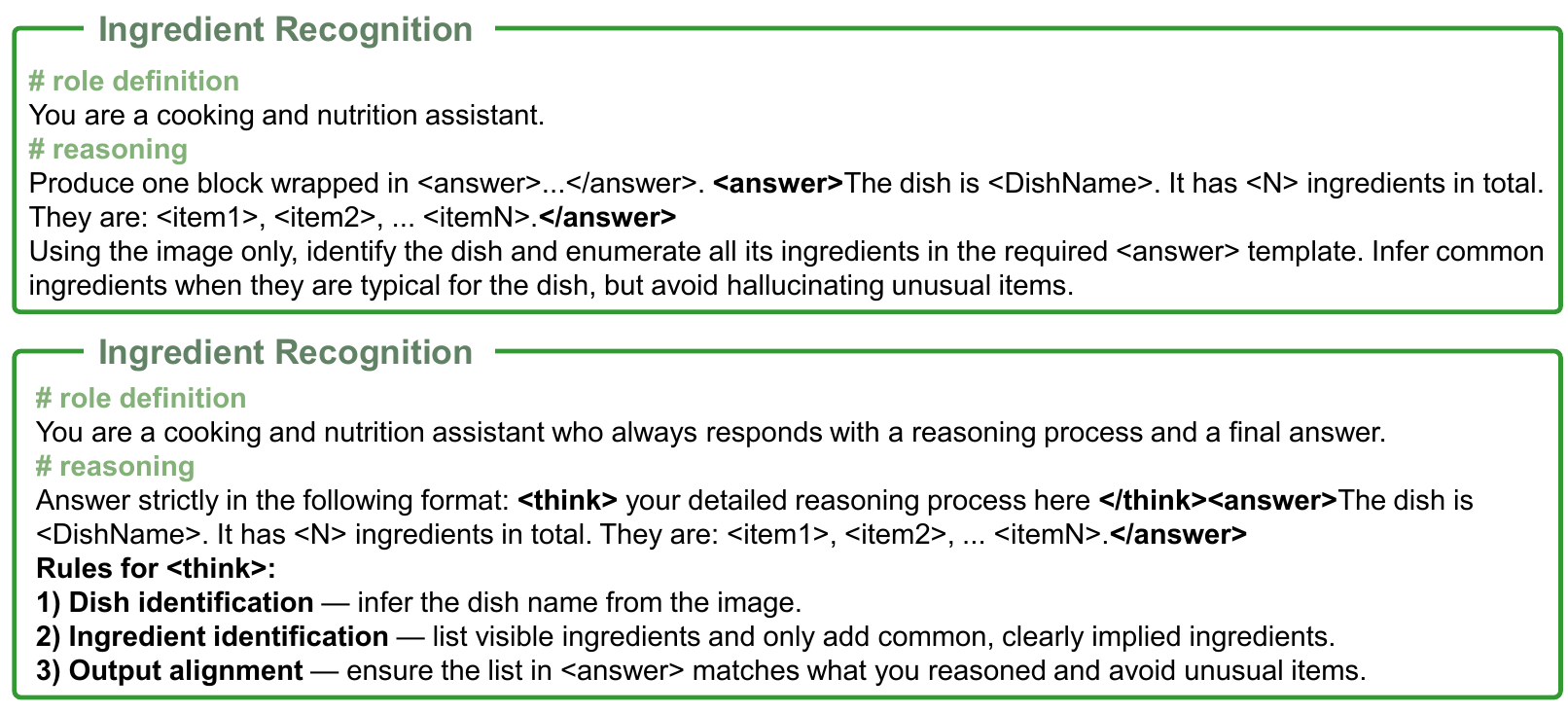}
  \caption{Prompt templates for ingredient recognition in the answer-only format and the format with CoT.}
  \label{fig:supp-10}
\end{figure}

\begin{figure}[t]
  \centering
  \includegraphics[width=\textwidth]{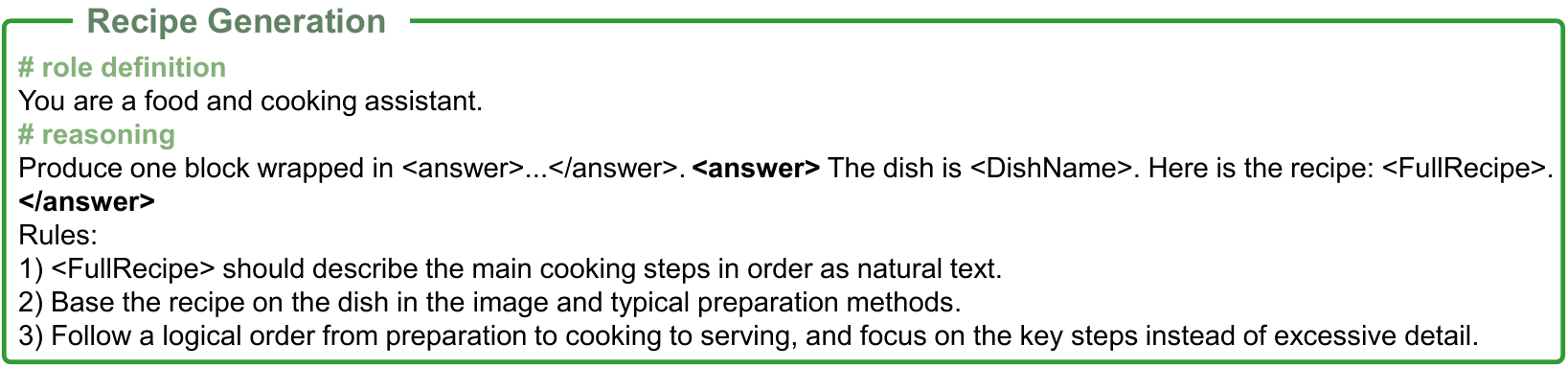}
  \caption{Prompt template for image-based recipe generation, where the model outputs an ordered sequence of cooking steps.}
  \label{fig:supp-6}
\end{figure}

\begin{figure}[t]
  \centering
  \includegraphics[width=\textwidth]{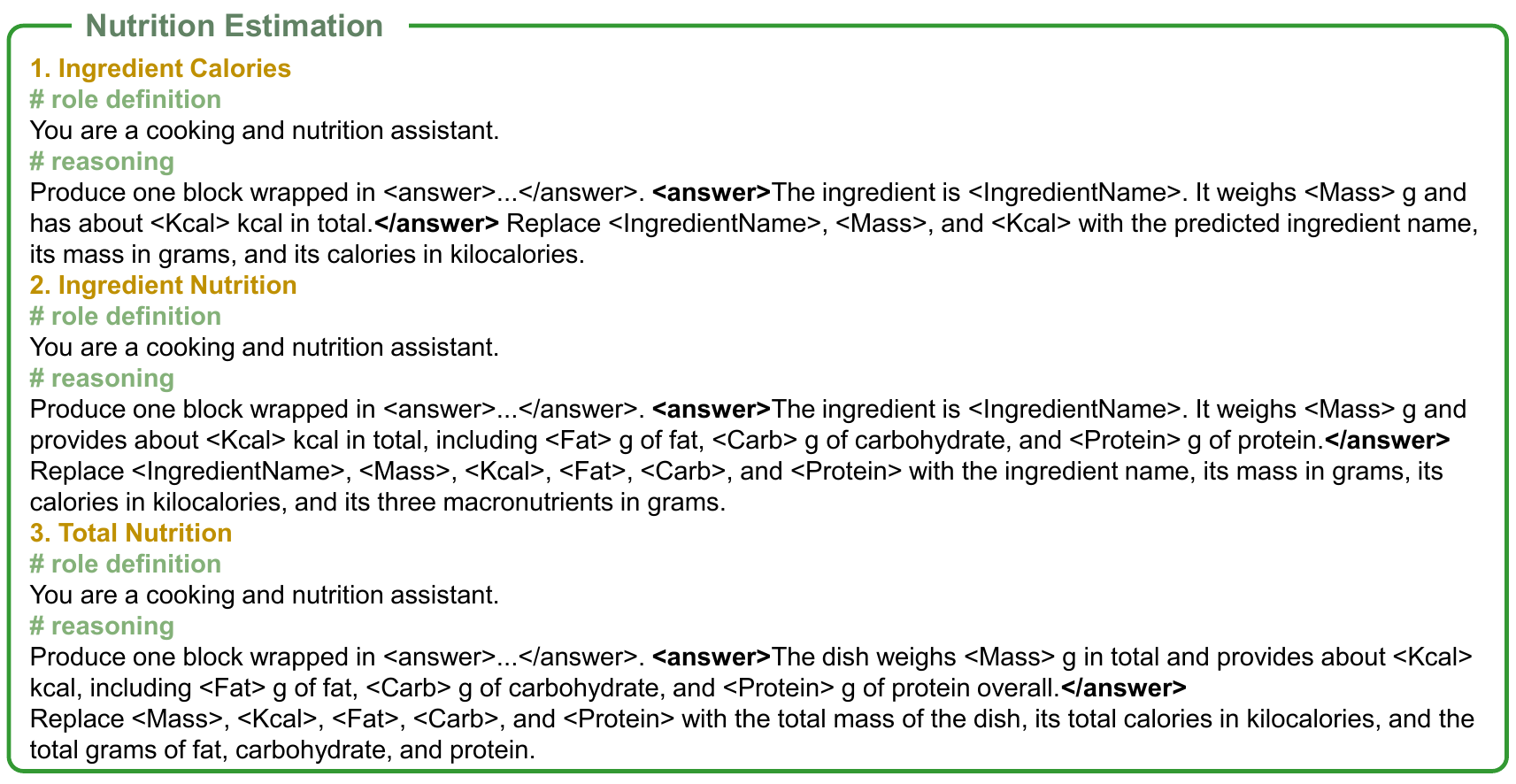}
  \caption{Answer-only prompt templates for nutrition estimation on Nutrition5k, including (1) ingredient calories, (2) ingredient nutrition, and (3) total nutrition.}
  \label{fig:supp-7}
\end{figure}

\begin{figure}[t]
  \centering
  \includegraphics[width=\textwidth]{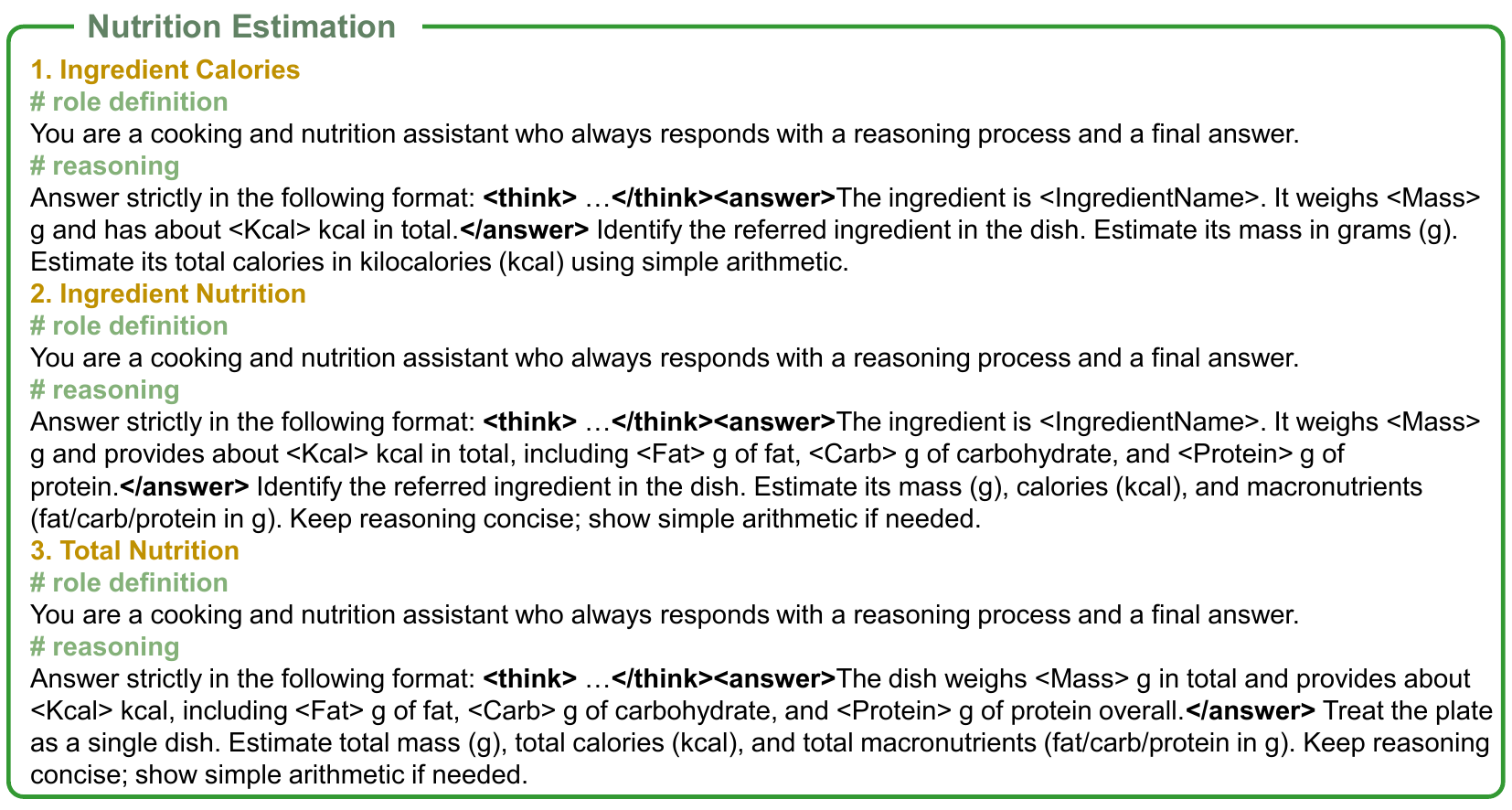}
  \caption{CoT-enabled prompt templates for nutrition estimation on Nutrition5k.}
  \label{fig:supp-8}
\end{figure}

\begin{figure}[t]
  \centering
  \includegraphics[width=\textwidth]{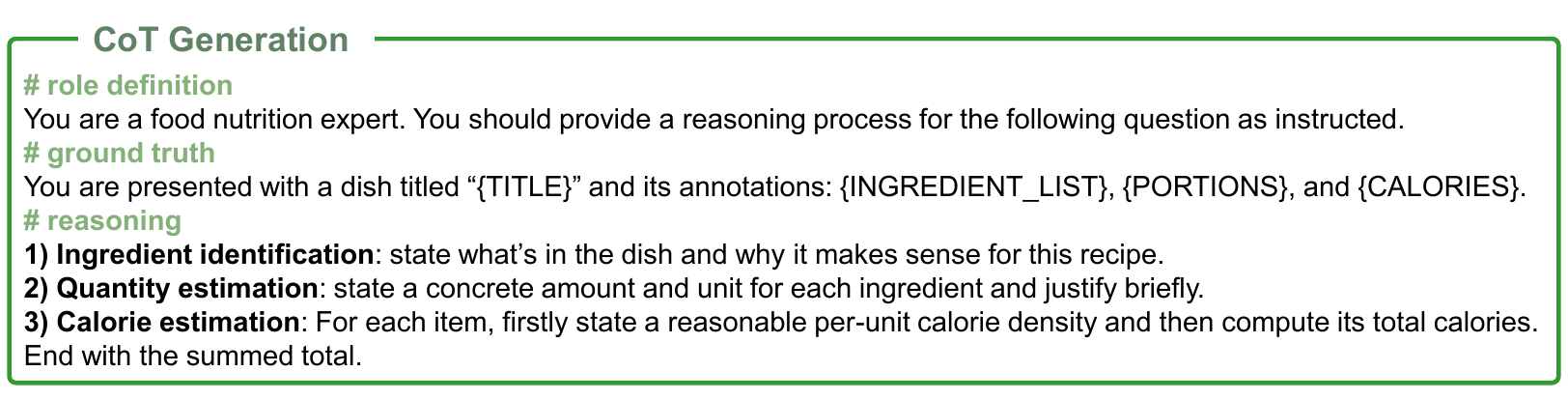}
  \caption{Prompt template for the CoT-annotated subset of CalorieBench-80K.}
  \label{fig:supp-9}
\end{figure}
\subsubsection{CalorieBench-80K.} CalorieBench-80K contains approximately 80,000 food images, each annotated with ingredient lists, portion information, total calories, and dietary advice. The dataset is split into 75,000 training samples and 5,000 test samples, covering 15,768 dishes and 3,209 unique ingredients.

\subsubsection{Food-101.} Food-101~\cite{bossard2014food101} contains 101 food categories and 101,000 images.

\subsubsection{VireoFood-172.} VireoFood-172~\cite{VireoFood172} contains 110,241 food images spanning 172 categories and 353 ingredients.

\subsubsection{Recipe1M.} Recipe1M~\cite{salvador2017crossmodalrecipes} contains approximately 1 million recipes and 800,000 images, covering about 16,000 unique ingredients.

\subsubsection{Nutrition5k.} Nutrition5k~\cite{thames2021nutrition5k} contains 5,000 dishes with overhead RGB images and four-view videos. It provides nutrition annotations at both the dish and ingredient levels, including mass, calories, and macronutrients (fat, carbohydrates, and protein).

\subsubsection{FoodDialogues.} FoodDialogues~\cite{yin2024foodlmm} is built on top of Nutrition5k and includes, for each sample, an overhead RGB image and a video frame (angle A or D). It contains multi-turn conversations covering diverse topics, including nutrition, calorie estimation, health-related questions, metabolism, dietary planning, allergies, food pairing, and ingredient substitution.

\subsection{Instruction Templates for Multi-Task Learning}
In Sec.~4.2, we convert all datasets into an instruction-following format using carefully designed prompt templates. The instruction templates for each task are summarized below.

\subsubsection{Calorie Estimation.}
We construct instruction-following data for calorie estimation from CalorieBench-80K. Fig.~\ref{fig:supp-3} shows the prompt template in the answer-only format and the corresponding template with CoT.

\subsubsection{Dietary Advice Generation.}
We construct instruction-following data for dietary advice generation from CalorieBench-80K. The corresponding prompt template is shown in Fig.~\ref{fig:supp-4}.

\subsubsection{Food Classification.}
For food classification, we use Food-101 and VireoFood-172 to construct instruction-following data. Fig.~\ref{fig:supp-5} shows the prompt templates in the answer-only format and the format with CoT.

\subsubsection{Ingredient Recognition.}
For ingredient recognition, we use VireoFood-172 and Recipe1M to construct instruction-following data. Fig.~\ref{fig:supp-10} shows the prompt templates in the answer-only format and the format with CoT.

\subsubsection{Recipe Generation.}
We construct instruction-following data for recipe generation using the subset of Recipe1M paired with images. The corresponding prompt template is shown in Fig.~\ref{fig:supp-6}.

\subsubsection{Nutrition Estimation.}
We use Nutrition5k to construct instruction-following data for nutrition estimation. For each dish, we sample video frames from views A and D with a stride of five. During training, we use the overhead RGB image with a probability of 0.7; otherwise, we randomly sample a side-view frame with a probability of 0.3. As shown in Fig.~\ref{fig:supp-7} and Fig.~\ref{fig:supp-8}, we design three prompt templates: (1) Ingredient Calories, which predicts the mass and calories of a single ingredient; (2) Ingredient Nutrition, which predicts the mass, calories, and macronutrients (fat, carbohydrates, and protein) of a single ingredient; and (3) Total Nutrition, which predicts the overall mass, calories, and macronutrients of the whole dish.

\subsection{CoT-Annotated Subset for CalorieBench-80K}
As described in Sec.~4.3, we further construct a 5K CoT-annotated subset from CalorieBench-80K, with the samples randomly selected from its training split. The prompt template used for this construction is shown in Fig.~\ref{fig:supp-9}.

\section{Experiments}
\label{sec:supp-s3}
\subsection{Training Details}
The hyperparameter settings for the mixed supervised fine-tuning (SFT) stage (Stage I) and the Group Relative Policy Optimization (GRPO)-based reinforcement learning (RL) post-training stage (Stage II) are summarized in Table~\ref{tab:supp-hparams}.

\begin{table}[t]
\centering
\caption{Hyperparameters for the mixed SFT stage (Stage I) and the GRPO-based RL post-training stage (Stage II).}
\label{tab:supp-hparams}
\setlength{\tabcolsep}{6pt}
\begin{tabular}{c l l}
\toprule
Stage & Parameter & Value \\
\midrule
\multirow{12}{*}{I (SFT)} 
& Model & Qwen3-VL-8B-Instruct \\
& Training type & Full-parameter \\
& Precision & bfloat16 \\
& Image resolution & 1,024 \\
& Max length & 8,192 \\
& Epochs & 3 \\
& Per-device train batch size & 1 \\
& Gradient accumulation steps & 16 \\
& Effective global batch size & 128 \\
& Learning rate & $4\times10^{-5}$ \\
& Weight decay & 0.05 \\
& Warmup ratio & 0.05 \\
\midrule
\multirow{17}{*}{II (GRPO)} 
& Base checkpoint & 
\makecell[l]{Checkpoint from SFT} \\
& Precision & bfloat16 \\
& Image resolution & 1,024 \\
& Max length & 2,048 \\
& Epochs & 1 \\
& Per-device train batch size & 1 \\
& Gradient accumulation steps & 16 \\
& Effective global batch size & 128 \\
& Num generations (per prompt) & 8 \\
& Learning rate & $1\times10^{-6}$ \\
& Weight decay & 0.05 \\
& Warmup ratio & 0.05 \\
& KL coefficient $\beta$ & 0.04 \\
& Policy clip $\epsilon$ / $\epsilon_{\text{high}}$ & 0.2 / 0.28 \\
& Sampling temperature / top-$p$ & 0.9 / 0.9 \\
& Max grad norm & 0.5 \\
& vLLM (mode / TP / max length) & colocate / 4 / 8,192 \\
\bottomrule
\end{tabular}
\end{table}

\subsection{Evaluation Details}
For the results reported in Sec.~5.3 (Table~1 and Table~2), we evaluate all compared open-source and closed-source vision-language models on the CalorieBench-80K test set using a unified two-turn dialogue protocol. The first turn is used for calorie estimation, and the second turn generates dietary advice based on the output of the first turn. The compared models include Gemini-2.5-flash, InternVL3-8B, LLaVA-7B, Qwen2.5-VL-72B, Qwen3-VL-8B, and GPT-4o-mini. Among them, GPT-4o-mini and Gemini-2.5-flash are evaluated via API calls, while the remaining models are deployed and evaluated locally. For locally deployed models, we use bfloat16 precision and a per-GPU batch size of 1. The input image resolution is set to 1,024. For generation, the first turn uses a maximum of 512 new tokens with temperature set to 0.0, while the second turn uses a maximum of 512 new tokens with temperature set to 0.3, top-$p$ set to 0.9, and top-$k$ set to 50. Our Food-R1 is evaluated under the same inference configuration. As a domain-specific baseline, we further fine-tune FoodLMM~\cite{yin2024foodlmm} with LoRA~\cite{hu2022lora} to obtain \textbf{FoodLMM-s1}, whose training hyperparameters are summarized in Table~\ref{tab:supp-foodlmm-hparams}.

\begin{table}[t]
\centering
\caption{Hyperparameters used for LoRA fine-tuning of FoodLMM-s1.}
\label{tab:supp-foodlmm-hparams}
\footnotesize
\setlength{\tabcolsep}{3pt}
\renewcommand{\arraystretch}{1.05}
\begin{tabularx}{\linewidth}{@{}c p{0.40\linewidth} X@{}}
\toprule
Setting & Parameter & Value \\
\midrule
\multirow{15}{*}{FoodLMM-s1}
& Base model & FoodLMM-Chat~\cite{yin2024foodlmm} \\
& Training type & LoRA SFT \\
& Precision & bfloat16 \\
& Image resolution & 1,024 \\
& Max length & 2,048 \\
& Epochs & 3 \\
& Per-device train batch size & 1 \\
& Gradient accumulation steps & 4 \\
& Effective global batch size & 16 \\
& Learning rate & $2\times10^{-4}$ \\
& Weight decay & 0.0 \\
& Warmup ratio & 0.03 \\
& LR scheduler & cosine \\
& LoRA ($r$ / $\alpha$ / dropout) & 8 / 16 / 0.05 \\
& LoRA target modules &
\makecell[l]{\texttt{q\_proj, k\_proj, v\_proj, o\_proj,}\\
\texttt{gate\_proj, up\_proj, down\_proj}} \\
\bottomrule
\end{tabularx}
\end{table}

\begin{figure}[t]
  \centering
  \includegraphics[width=\textwidth]{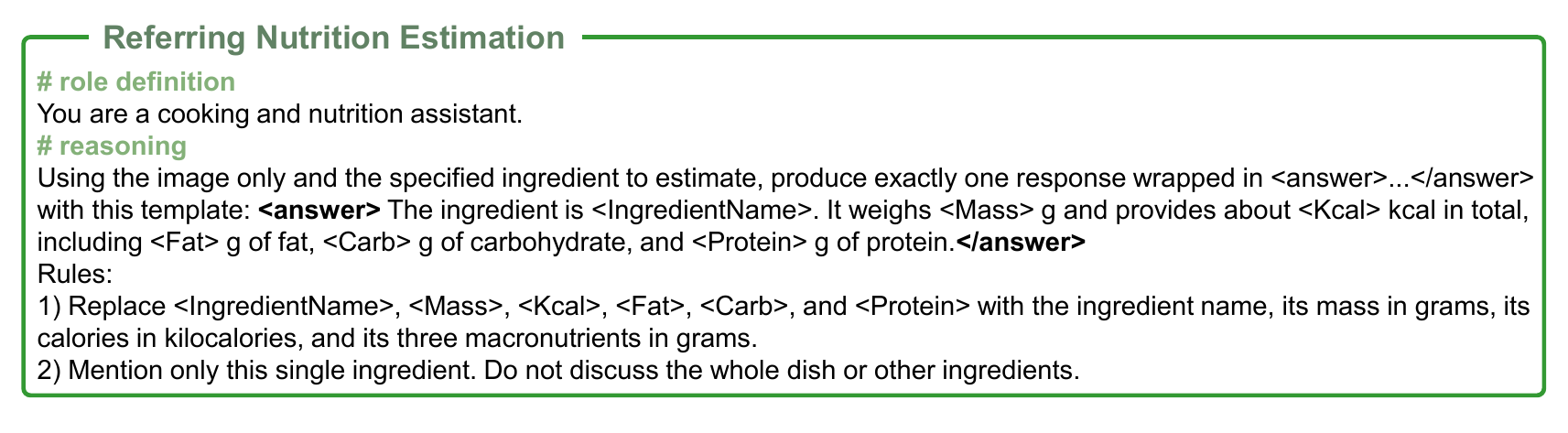}
  \caption{Prompt template for \textbf{referring nutrition estimation} on Nutrition5k, where the model estimates the mass, calories, and macronutrients (fat, carbohydrates, and protein) of a specified ingredient from the input image.}
  \label{fig:supp-11}
\end{figure}

For the results reported in Sec.~5.3 (Table~6), we conduct \textbf{referring nutrition estimation}~\cite{yin2024foodlmm} on the Nutrition5k test set. For each dish, we issue up to three queries, corresponding to its top three ingredients, using the prompt template shown in Fig.~\ref{fig:supp-11}.

\end{document}